\documentclass[10pt,twocolumn,letterpaper]{article}

\usepackage{cvpr}
\usepackage{times}
\usepackage{epsfig}
\usepackage{graphicx}
\usepackage{amsmath}
\usepackage{amssymb}
\usepackage{algorithm} 
\usepackage{algorithmic}  
\usepackage[algo2e]{algorithm2e}
\usepackage{breqn}

\usepackage[pagebackref=true,breaklinks=true,letterpaper=true,colorlinks,bookmarks=false]{hyperref}

\cvprfinalcopy 

\def\cvprPaperID{3569} 

\ifcvprfinal\fi
\begin{document}

\title{Generating a Fusion Image: One's \textit{Identity} and Another's \textit{Shape}}

\author{Donggyu Joo$^*$ ~~~~~~~~~~~~~Doyeon Kim$^*$ ~~~~~~~~~~~~~ Junmo Kim
\\
School of Electrical Engineering, KAIST, South Korea\\
{\tt\small \{jdg105, doyeon\_kim, junmo.kim\}@kaist.ac.kr}
\\
}

\maketitle

\begin{abstract}
Generating a novel image by manipulating two input images is an interesting research problem in the study of generative adversarial networks (GANs). We propose a new GAN-based network that generates a fusion image with the \textit{identity} of input image $x$ and the \textit{shape} of input image $y$. Our network can simultaneously train on more than two image datasets in an unsupervised manner. We define an identity loss $L_I$ to catch the identity of image $x$ and a shape loss $L_S$ to get the shape of $y$. In addition, we propose a novel training method called Min-Patch training to focus the generator on crucial parts of an image, rather than its entirety. We show qualitative results on the VGG Youtube Pose dataset, Eye dataset (MPIIGaze and UnityEyes), and the Photo--Sketch--Cartoon dataset.

\end{abstract}


\section{Introduction}

\vspace*{-3mm}
\let\thefootnote\relax\footnote{*These two authors contributed equally}

\begin{figure}[t]
\begin{center}
   \includegraphics[width=1.0\linewidth]{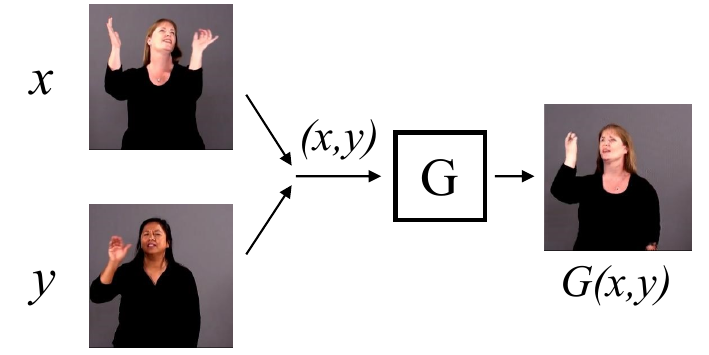}
\end{center}
\vspace*{-3mm}
   \caption{Task of our network. There are two input images $x$ and $y$. Generated output $G(x,y)$ is combined image which has $x$'s \textit{identity} and $y$'s \textit{shape}.}
	\label{fig: mainidea}
	
\vspace*{-3mm}

\end{figure}

Transforming an object or person to a desired shape is a well-used technique in real world. For example, computer graphics have made it possible to display scenes on screen that would be difficult to implement physically, like making a person pose in a posture that was not actually taken. The approach we use in this study aims to produce a fusion image that combines one image's \textit{identity} with another image's \textit{shape}, a task that is demonstrated in Figure~\ref{fig: mainidea}.

In machine learning, generative models such as variational autoencoders (VAEs)~\cite{kingma2013auto} and restricted boltzmann machine (RBM)~\cite{smolensky1986information} have been able to generate new images that follow the input data distribution. Furthermore, the development of generative adversarial networks (GANs)~\cite{goodfellow2014generative} has led to revival of generative models. GANs have produced realistic images of human face, furniture, or scene that are difficult to distinguish from real images~\cite{Radford2015UnsupervisedRL}. Consequently, GANs have been used to create target images in various fields such as text-to-image conversion, face completion, and super-resolution imaging~\cite{Ledig_2017_CVPR, li2017generative, reed2016generative, Zhang_2017_ICCV}. Especially, image to image translation, which changes the characteristics of an image to those the user wants, succeeded in translating the images~\cite{Isola_2017_CVPR}. Such translation has produced reasonable results from both paired and unpaired data in tasks including transforming an apple to an orange, a photo to a painting, and a winter scene to a summer scene~\cite{zhu2017unpaired}. Our task is also a kind of image to image translation. We can change the identity of an image within a desired shape. However, since existing image translation learns the mapping function from one set to the other without explicit loss functions for shape matching, there is a possibility that shape may not remain the same. Therefore, we need another framework to deal with both \textit{identity} and \textit{shape} effectively.

In this paper, we propose a framework for creating a result image that follows the shape of one input image and possesses the identity of another image. Having a label for a result image, (\ie, in a supervised setting) can make it easier to generate a corresponding image. However, since such ground truth label images for new combinations of \textit{identities} and \textit{shapes} are usually unavailable, we need to create our result images without the label images.
\begin{figure*}[t]
\begin{center}
   \includegraphics[width=0.9\linewidth]{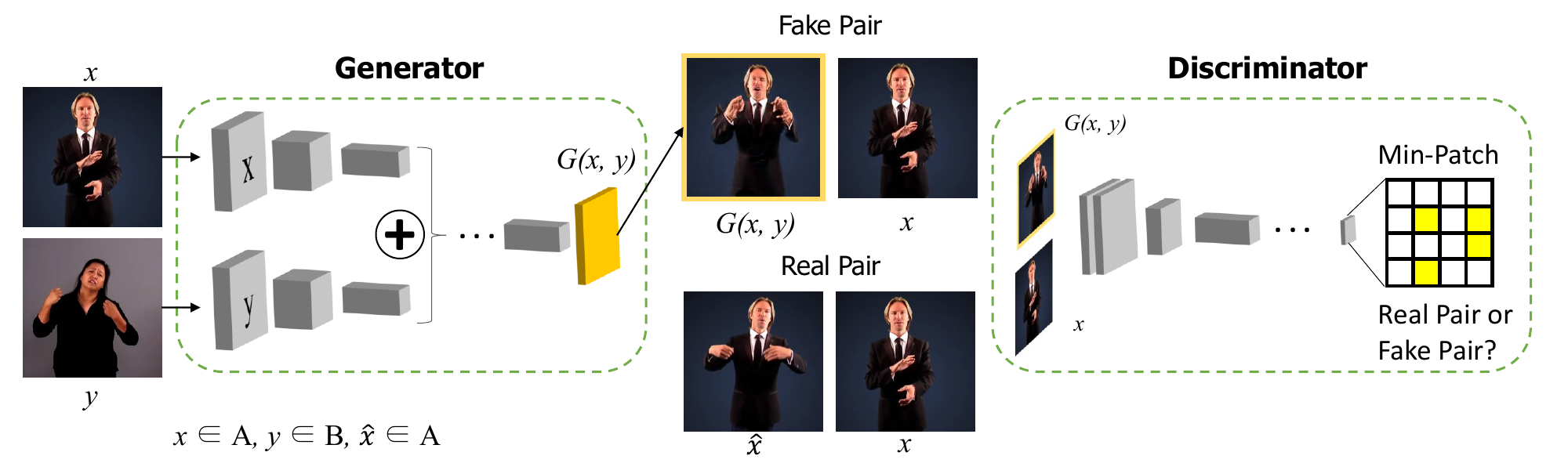}
\end{center}
\vspace*{-2mm}
   \caption{Architecture of our work: Input images $x$ and $y$ come into the generator together. Then, the discriminator distinguishes between real image pair and fake pair that consists of a generated image and a real image. At the final layer of the discriminator, we use Min-Patch training to focus on more important parts of an image. }
\vspace*{-3mm}
\label{fig:architecture}
\end{figure*}

We demonstrate that FusionGAN achieves our goal by showing result images that follow the identity of one input image and the shape of another. From YouTube Pose dataset~\cite{Charles16}, we generate an image with one person's identity and another person's pose. We generate realistic eye images that follow the shape of synthetic eye images in the Eye dataset. Finally, we transform a person's face into a different style using Photo--Sketch--Cartoon sets. 

\section{Related Work}

GANs proposed by Goodfellow \etal~\cite{goodfellow2014generative}, generated a more realistic image than the previous studies. It contains two components: a discriminator, which distinguishes real images from fake ones, competes with a generator, which generates the fake images that look like real. In their adversarial training procedure, the generator generates outputs whose distribution gets closer to the real data distribution. GANs have been applied to various tasks, such as text-to-image conversion, super-resolution, or data augmentation~\cite{Ledig_2017_CVPR,reed2016generative, sixt2016rendergan}.

Isola \etal~\cite{Isola_2017_CVPR} successfully performed image translation on varied datasets by using the GAN-based ``pix2pix'' framework, which allows a generator to learn the mapping function between two paired images. In their framework, the fake pair consists of the input and the generated output. And the discriminator tries to distinguish between the real and fake pairs. However, pix2pix framework needs paired data, which is difficult to obtain. CycleGAN ~\cite{zhu2017unpaired} enabled unpaired image to image translation, using the idea of cycle consistency. For two sets of images, $X$ and $Y$, they trained two translators $G : X \rightarrow Y$ and $F : Y \rightarrow X$ so that $F(G(x))$ becomes the same as original image $x$. However, CycleGAN requires more than one model to deal with multiple sets.

There have also been many attempts to change an image's attributes, rather than the whole image. InfoGAN~\cite{Chen2016InfoGANIR} sets the loss to maximize mutual information between code vector $c$ and generated outputs, leading to a network that can learn image attributes. The DiscoGAN~\cite{pmlr-v70-kim17a} framework is similar to that of CycleGAN, but its target task is to manipulate attributes of image such as hair color or sunglasses. Deep feature interpolation (DFI)~\cite{Upchurch_2017_CVPR} achieved high-resolution semantic transformations by linear interpolation of features obtained from a pre-trained convolutional neural networks.

The goal of $PG^2$~\cite{ma2017pose} and visual analogy making is similar to ours, in that they transform a query image to a desired appearance. $PG^2$ manipulated a reference image of a person to a guided pose using human body keypoints. They annotated the keypoints with existing pose estimation techniques and used them in training. Reed \etal~\cite{reed2015deep} proposed a method for visual analogy making whose task is to change a query image to a target that constitutes a related pair of images. 

\begin{figure*}[t]
\begin{center}
   \includegraphics[width=0.9\linewidth]{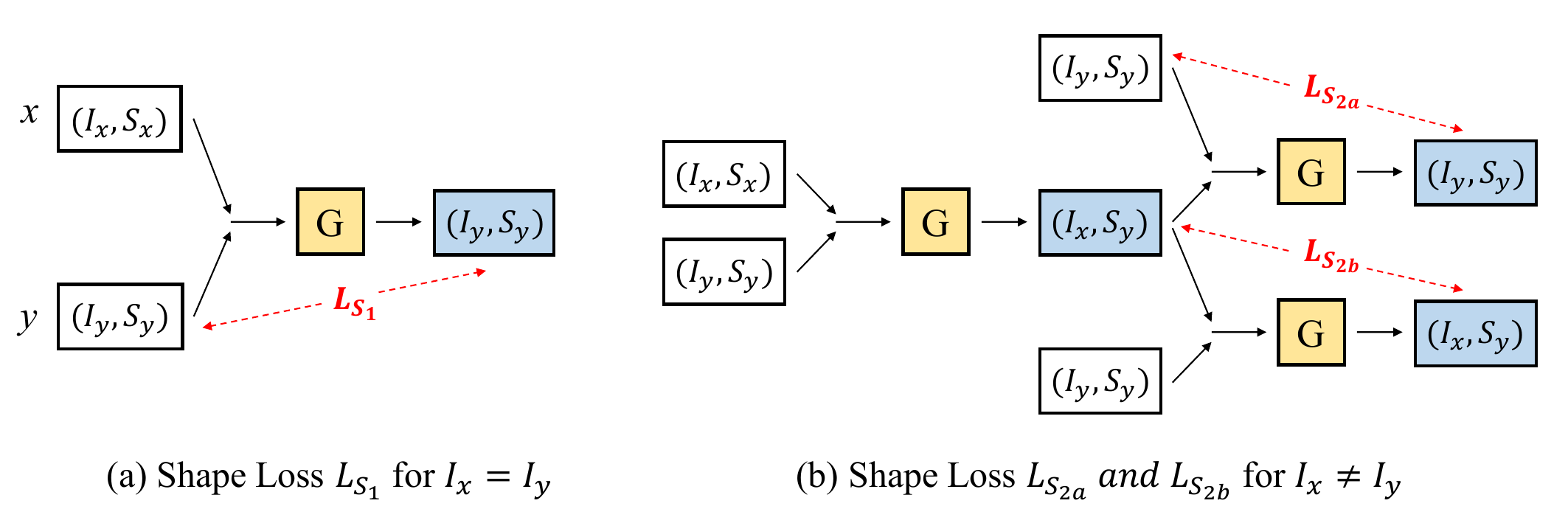}
\end{center}
\vspace*{-3mm}
   \caption{Illustration of \textit{Shape Loss}. (a) Shape loss $L_{S_1}$ is defined when $I_x = I_y$ (b) Shape loss $L_{S_{2a}}$ and $L_{S_{2b}}$ are defined when $I_x \neq I_y$. To achieve our goal, these shape losses should be minimized. In the figure, generator $G$ is only one model (same weights), and blue images are the generated output images from the generator}
\label{fig:shapeloss}
\vspace*{-3mm}
\end{figure*}

\section{Methods}

The goal of our work is to learn a mapping function that generates a fusion image from two input images given from multiple unlabeled sets. By getting the \textit{identity} from the first input image $x$ and the \textit{shape} from the second input image $y$, we can generate a combined output image. We express image $x$, with \textit{identity} $I_x$ and \textit{shape} $S_x$, as $x=(I_x, S_x)$. When our network has two input images $x=(I_x, S_x)$ and $y=(I_y, S_y)$, our goal is to generate the following new fusion image:
\begin{equation}
G(x=(I_x, S_x), y=(I_y, S_y))=(I_x, S_y)
\end{equation}

Thus, the output is a fusion image that has the same \textit{identity} as $x$, and the same \textit{shape} as $y$. When two inputs are given, generator $G$ can automatically get the \textit{identity} from $x$, and \textit{shape} from $y$ (Figure~\ref{fig: mainidea}). Our network is not limited to the transition between two image sets, but can be used for multiple unlabeled image sets, unlike previous works CycleGAN~\cite{zhu2017unpaired} and DiscoGAN~\cite{pmlr-v70-kim17a}. The identity and shape we mention here can be any characteristics, depending on the image sets and task. In general, \textit{identity} can be seen as a set-level characteristic which all images in a set share, and \textit{shape} can be seen as an instance-level characteristic that can distinguish every single image within the set. For example, in the case of the VGG YouTube Pose dataset~\cite{Charles16}, where each set consists of multiple images of one person with multiple poses, \emph{identity} of an image is indeed the who the person is in the image and \textit{shape} is the person's pose. We demonstrate various kinds of set definition in Section \ref{sec:experiment}.

\subsection{Identity Loss $L_I$}
To get the identity of image $x$, we need to make the distribution of the output image similar to the distribution for the set of images to which the first input image $x$ belongs. To do so, we used a pair discriminator $D$ that discriminates whether the input pair is a real pair or fake pair. We express identity loss as:
\begin{multline}
L_I(G, D)=\mathbb{E}_{x,\hat{x}\sim p_{data}(x)}[\log D(x, \hat{x})]\\
 +\mathbb{E}_{x\sim p_{data}(x), y\sim p_{data}(y)}[\log (1-D(x, G(x, y)))],
\end{multline}
where $x$ and $y$ are the two network inputs, and $\hat{x}$ is another image having the same identity as $x$, \ie, $I_{x}=I_{\hat{x}}$. $x\sim p_{data}(x)$ and $y\sim p_{data}(y)$ represents the data distribution. This identity loss includes $G$ and $D$, and we apply adversarial training to achieve the goal. $G$ tries to generate desired image $G(x,y)\sim p_{data} (x)$ while $D$ aims to distinguish between a real pair  $(x, \hat{x})$ and a fake pair $(x,G(x,y))$. That is, $D$ tries to maximize $L_I$ and $G$ aims to minimize it. After adversarial training, output $G(x,y)$ is generated to have the same identity as input $x$. The whole architecture of our network is shown in Figure~\ref{fig:architecture}. The generated output is input to the discriminator as a pair with input $x$.

We use $L2$ loss instead of the negative log likelihood. Now, $L_I(G,D)$ is replaced by the following equation. We train the $G$ to maximize $L_I$ and $D$ to minimize $L_I$. Replaced identity loss $L_I$ is 
\begin{multline}
L_I(G, D)=\mathbb{E}_{x,\hat{x}\sim p_{data}(x)}[||1-D(x, \hat{x})||_2]\\
 +\mathbb{E}_{x\sim p_{data}(x), y\sim p_{data}(y)}[||D(x, G(x, y))||_2].
\end{multline}

\subsection{Shape Loss $L_S$}

If we train the network by using only the objective $L_I$, we get a random image with an identity of $x$ that is independent of $y$. Therefore, we need to design another objective that extracts the \textit{shape} from the second input image $y$. Since we are using unlabeled multiple image sets that have no shape label, it is difficult to define what the resulting image $(I_x, S_y)$ should be. 
CycleGAN~\cite{zhu2017unpaired} and DiscoGAN~\cite{pmlr-v70-kim17a} use a loss that input image must return to the original input image after passing through two consecutive translations.
Shape preserving ability in those works is achieved by such a loss that encourages returning the image back to its original input. Since it is not directly trained to retain shape and it may be possible to learn an inverse mapping of the first translation without the ability to preserve shape perfectly, there is a risk of under-preservation of shape.

We propose another effective way of preserving shapes by novel design of shape loss functions. What if two input images have the same identity? Then, the second input image $y$ should be the output image $G(x,y)$. Using this idea, we designed the following loss to extract the shape from the second image. Then the network can generate a target image even though the ground truth does not exist. A simple illustration of shape losses is described in Figure~\ref{fig:shapeloss}. For two input images $x$ and $y$ with the same identity, \ie, $I_{x}=I_{y}$, shape loss $L_{S_1}(G)$ is defined as:
\begin{equation}
L_{S_1}(G)=\mathbb{E}_{x\sim p_{data}(x), y\sim p_{data}(y)}[||y - G(x, y)||_1].
\end{equation}

Since this is a simple $L1$ loss, if we only use $L_{S_1}(G)$, the generator's learning may focus only on getting the second image. Therefore, we added two more losses that have the same concept as above, but try to get information from the first input of the generator. In the case of a human, let's say that person $x$ and $y$ belong to sets A and B, respectively. If $G(x,y)$ comes out as we want it, then $G(x,y)$ would be a figure of $x$ posing as $y$. So what happens if the resulting image passes through the generator again with $y$? If $G(x,y)$ goes with $y$ to pose like $y$, then $G(x,y)$ should come out because it already has a pose of $y$. And if $y$ goes into resembling a pose of $G(x,y)$, then $y$ should come out. This helps to learn \textit{shape} while preventing the generator from being biased to one side of the input. For the following losses, two inputs do not need to have the same identity. For two input images $x$ and $y$ with different identities, \ie, $I_x\neq I_y$, additional shape loss is defined as:
\begin{equation}
\small
    L_{S_{2a}}(G)=\mathbb{E}_{x\sim p_{data}(x), y\sim p_{data}(y)}[||y - G(y, G(x, y))||_1],
\end{equation} 
\vspace{-8mm}
\begin{multline}
\small
    L_{S_{2b}}(G)=\mathbb{E}_{x\sim p_{data}(x), y\sim p_{data}(y)}\\
    [||G(x, y)-G(G(x, y), y)||_1].
\end{multline} 

Shape loss must be minimized to achieve our goal of the generator producing the shape of the second input. Then, the overall shape loss $L_S$ becomes the sum of these $L_{S_1}$, $L_{S_{2a}}$, and $L_{S_{2b}}$:

\begin{equation}
L_S(G)=L_{S_1} + \alpha (L_{S_{2a}}+L_{S_{2b}}).
\end{equation}

Thus, the overall loss function used in our paper is:
\begin{equation}
L(G,D)=L_I+\beta L_S
\end{equation}
where $\alpha$ and $\beta$ are hyper-parameters.

\subsection{Training Algorithm}

Before training, we prepare unlabeled multiple image sets $A, B, C, \cdots$. All the images in each set share the same set-level characteristic which we call \emph{identity}, and each image in the set has an individual instance-level characteristic which we call \textit{shape}. Therefore, $I_x=I_y$ if image $x$ and image $y$ belong to the same set. 

Our training algorithm follows the basic flow of GANs. We train both the generator and discriminator adversarially to generate a realistic desired image. Since we need different types of input pairs for two losses $L_I$ and $L_S$, we need two separate steps. The training procedure of our FusionGAN is described in Algorithm~\ref{algorithm for training}.

\begin{figure}[t]
\begin{center}
   \includegraphics[width=1.00\linewidth]{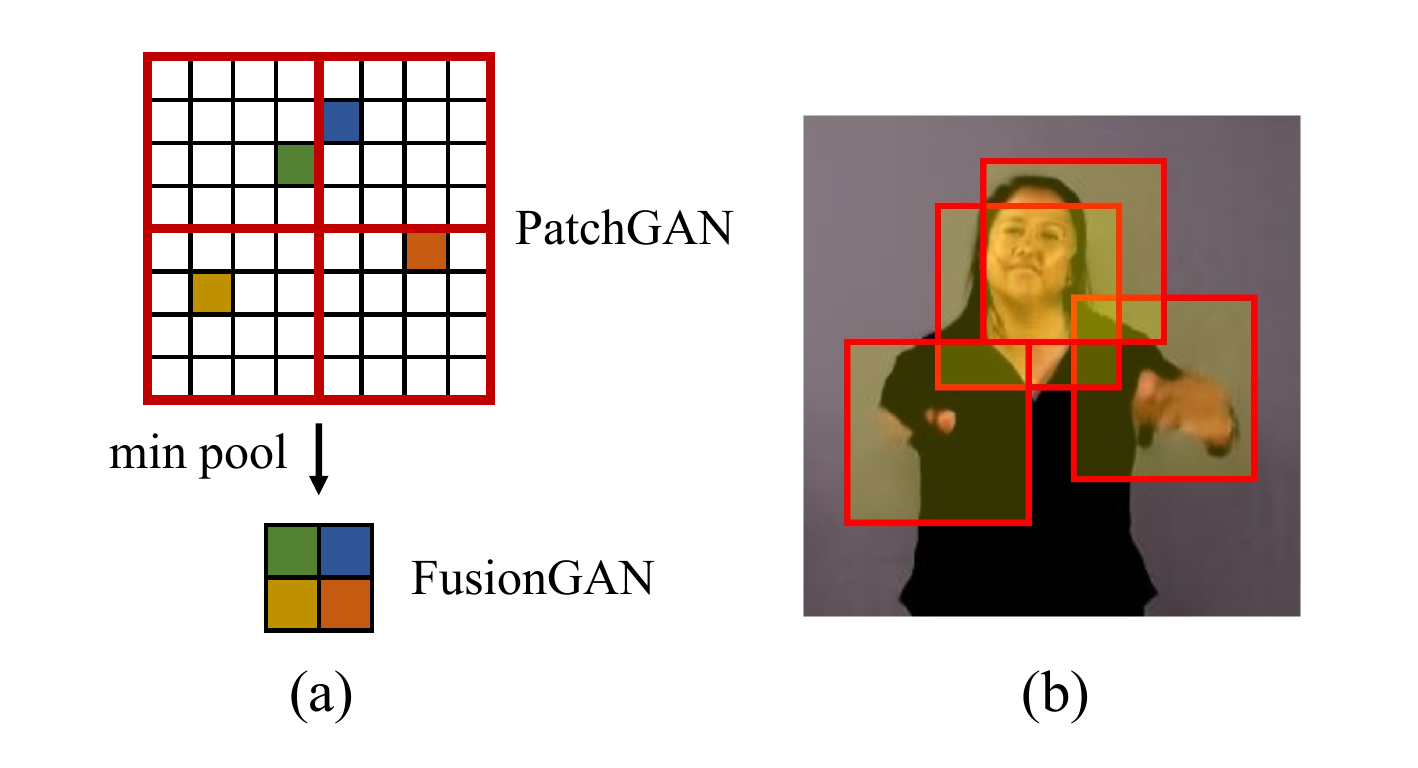}
\end{center}
\vspace*{-3mm}
   \caption{Illustration of \textit{Min-Patch training}. Minimum pooling is applied to the final output of discriminator when our generator is trained. When training discriminators, we use PatchGAN without min-pool. (a) Minimum pooling is applied to the last feature map of the discriminator. (b) Corresponding receptive field of the input image.}
\vspace*{-3mm}
\label{fig:min_pool}
\end{figure}

\begin{figure*}[t]
\begin{center}
   \includegraphics[width=0.8\linewidth]{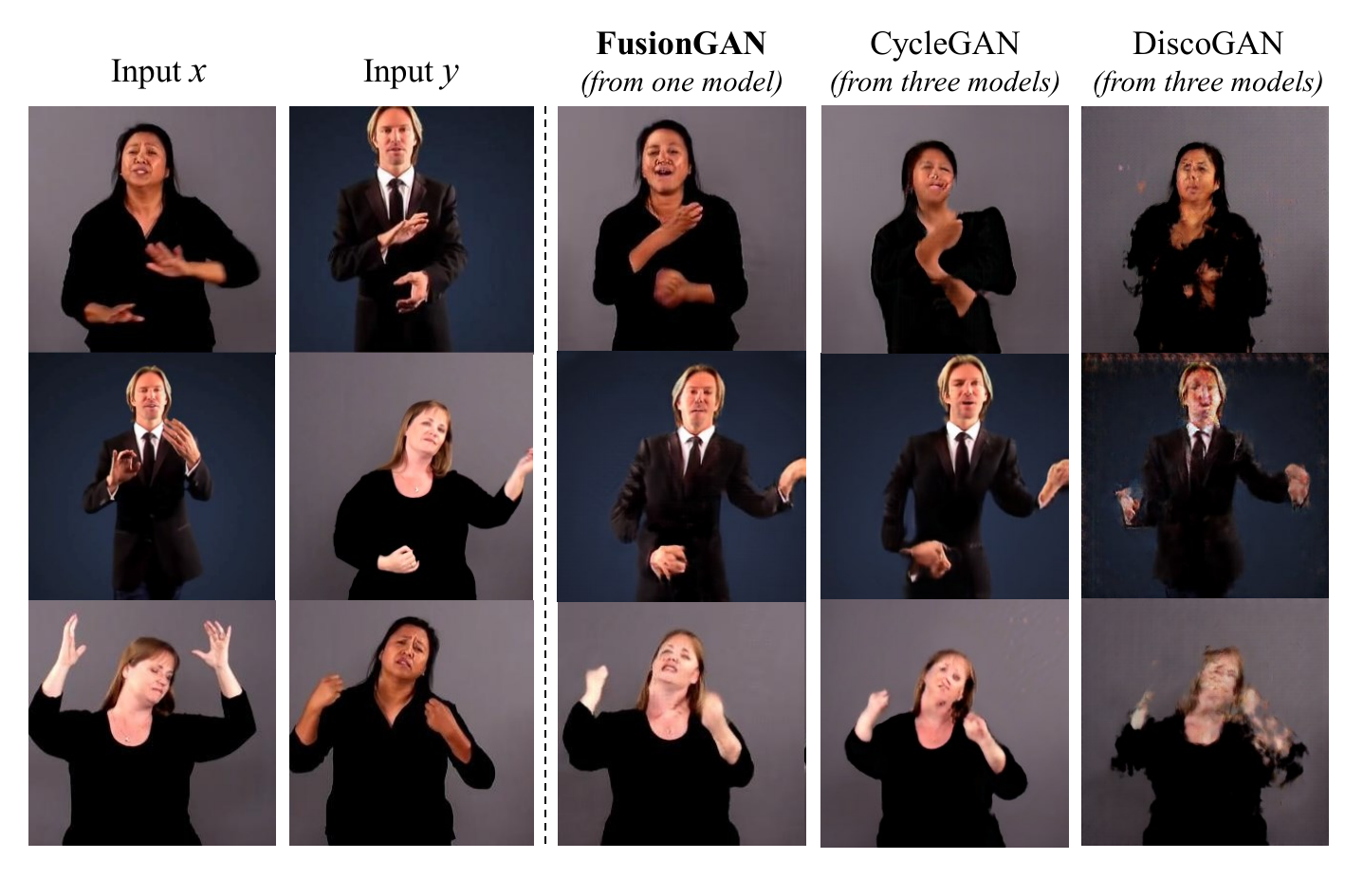}
\end{center}
\vspace*{-7mm}
   \caption{Results on the YouTube Pose Dataset~\cite{Charles16}. From left to right: input $x$, input $y$, FusionGAN, CycleGAN~\cite{zhu2017unpaired}, and DiscoGAN~\cite{pmlr-v70-kim17a}. We used only one generator. CycleGAN and DiscoGAN are both trained for three pairs of identities. For the CycleGAN and DiscoGAN, outputs are generated as $G_1(y), G_2(y)$ and $G_3(y)$ using three generators. They need three generators for this experiment.
}
\vspace*{-3mm}
\label{fig:Youtube}
\end{figure*}
\subsection{Min-Patch Training}

The original discriminator in GANs produces a single output that observes whole images at once. Recently, PatchGAN~\cite{Ledig_2017_CVPR, li2016precomputed, zhu2017unpaired} has been used to focus on every partial patches of an image. PatchGAN produces an output feature map $D(x)\in \mathbb{R}^{h\times w}$ instead of a single output value and uses the sum of all output values in that feature map. It discriminates an image by observing all partial patches comprehensively. However, when humans observe an image, we do not focus on every small part of it. Even if most parts seem realistic, if a small part of the image is strange, we feel that the image is a fake. Therefore, when we generate realistic images, it is better to concentrate on the strangest parts of the image. 

\begin{algorithm}[H]
 \KwData{Multiple image sets $A, B, C, \cdots$(\# sets $\ge 2$). Each image set has images with the same identity, and each image has individual shapes.}
 \textbf{Initialize:} All the settings include learning rates, optimization method, training iteration, initial weights, \etc.
 \vspace*{1mm}
 
 \While{not converged}{
  
  \vspace*{0.1cm} \textbf{(I) Get input image $x$ and $y$ each from two different randomly chosen sets, \ie $I_x\neq I_y$:}
  
  \vspace*{0.1cm} \hspace*{0.5cm} Update the $G$ $\leftarrow$ maximizes \textit{Identity loss} $L_I$\\
  \hspace*{2cm} (\textit{Min-Patch Training} is applied)
  
  \vspace*{0.1cm} \hspace*{0.5cm} Update the $D$ $\leftarrow$ minimizes \textit{Identity loss} $L_I$
  
  \vspace*{0.1cm} \hspace*{0.5cm} Update the $G$ $\leftarrow$ minimizes \textit{Shape loss} $L_{s_{2a}}, L_{s_{2b}}$
  \vspace*{0.1cm}
   
  \vspace*{0.1cm} \textbf{(II) Get input image $x$ and $y$ from one randomly chosen set, \ie $I_{x}=I_{y}$:}
  
  \vspace*{0.1cm} \hspace*{0.5cm} Update the $G$ $\leftarrow$ minimizes \textit{Shape loss} $L_{S_1}$
 }
 \caption{Training Algorithm for FusionGAN}
 \label{algorithm for training}
\end{algorithm}

We suggest \textit{Min-Patch training}, which uses minimum pooling at the last part of the PatchGAN discriminator when we train $G$ adversarially. $G$ tries to maximizes the objective $L_I$ after passing the minimum pooling because the lower discriminator output implies fake patch. So the generator $G$ is trying to maximize $D$ values after minimum pooling solving one more max-min problem as an inner loop. Min-Patch training is described in Figure~\ref{fig:min_pool}. PatchGAN training looks at every part of an input image ($8\times 8$ in the figure); however, Min-Patch training only focuses on the most important part of fake images ($2\times 2$ in the figure) when training $G$. It helps to generate realistic images because $G$ is trained so that there are less patches where it looks strange.

\section{Experiment}
\label{sec:experiment}

To confirm our method's successful result, we conducted experiments using various datasets. We compared our result with other unlabeled generative networks in various settings.

\textbf{Implementation details:} The generator network has two inputs $x$ and $y$. Each $x$ and $y$ passes separate three convolutional layers and two residual blocks, and then two output feature maps are concatenated. After passing through the two more residual blocks and two deconvolutional layers, it generates an output image. The generator has U-Net~\cite{ronneberger2015u} like structure to preserve low-level information well. 

For the Section \ref{sec: YouTube} and Section \ref{sec:cartoon}, the discriminator outputs $32\times 32$ size of feature maps. Our Min-Patch training uses $8\times 8$ min pooling, so generator uses output $4\times 4$ size of feature map for training the $L_I$. For the Section \ref{sec: eye}, the discriminator outputs $15\times 9$ size of feature maps. Min-patch training uses $3\times 3$ min pooling for this case, so generator uses output $5\times 3$ size of feature map for $L_I$.

\subsection{VGG YouTube Pose dataset}
\label{sec: YouTube}

We trained our network on a subset of VGG YouTube Pose dataset~\cite{Charles16}. We made three image sets $A$, $B$, and $C$ by capturing frames of three videos in YouTube Pose dataset. Each image set has all captured images of only one person with various poses. Thus, these three image sets contain various pose images of three people. Here, \textit{identity} refers to each person, and \textit{shape} refers to the person's pose. All images are generated in high-resolution $(256\times 256)$. For each set, $638$, $663$, and $686$ images were used for training. For testing, $157$ images which are disjoint from training sets were used.

\textbf{Qualitative Results:} Figure~\ref{fig:Youtube} shows our qualitative results on the YouTube Pose dataset. We compared our results with those of well-known unlabeled set translation methods, CycleGAN~\cite{zhu2017unpaired} and DiscoGAN~\cite{pmlr-v70-kim17a}. Our network generates output as $G(x,y)$. CycleGAN and DiscoGAN generate outputs as $G_1(y), G_2(y)$ and $G_3(y)$ for three kinds of set transitions where $G_1:A\rightarrow B$, $G_2:B\rightarrow C$, and $G_3:C\rightarrow A$. Our network generated more convincing results close to the desired image that has the identity of $x$ and the shape of $y$. Details such as arms and faces are generated well. It is to our advantage that our method learns shapes from unlabeled sets.

Moreover, CycleGAN and DiscoGAN required several separate generators for each pair of identities to handle various transformation in multiple sets. If there are three identities, they need $6$ generators to cover all the combinations. However, our network can handle several kinds of image sets using only one model. If there are more identities, this difference is amplified.

\begin{figure*}[t]
\begin{center}
   \includegraphics[width=0.87\linewidth]{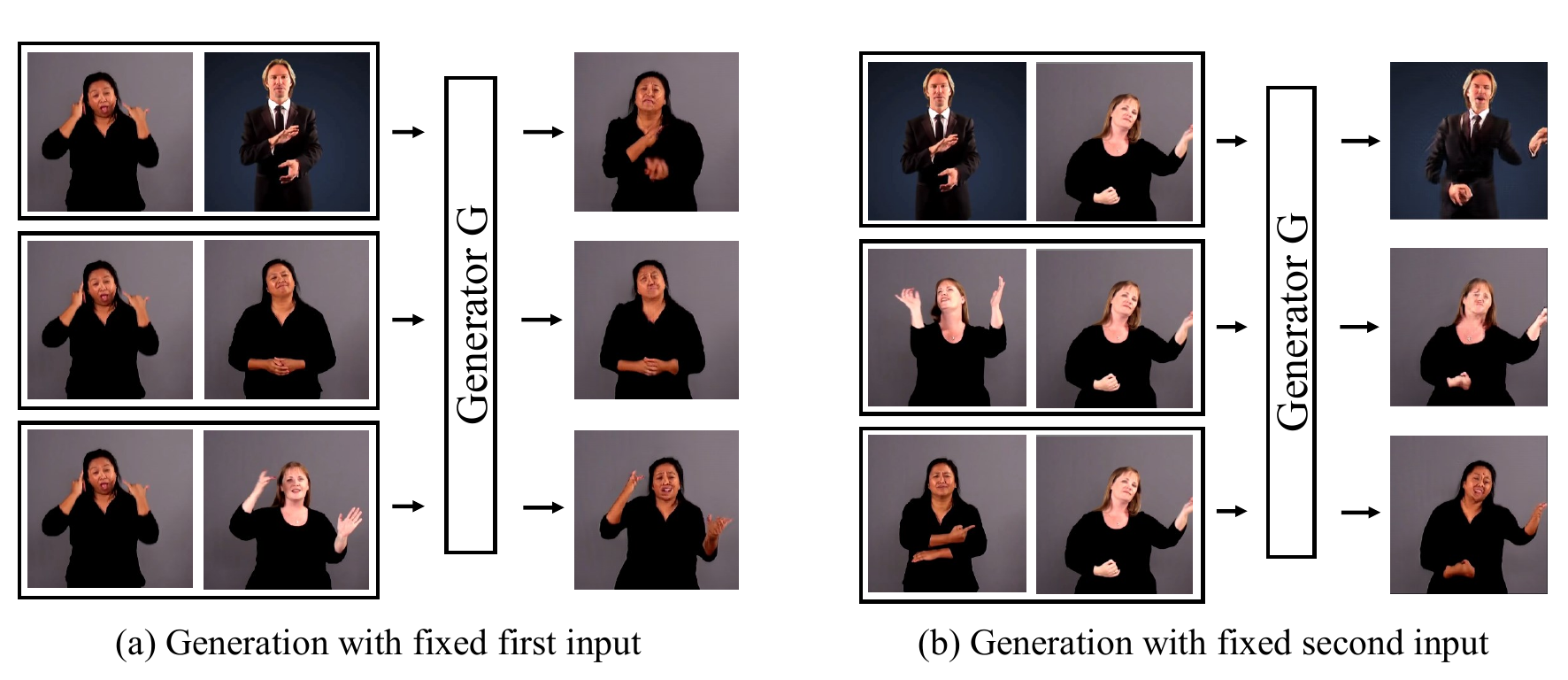}
\end{center}
   \caption{An efficient generation with FusionGAN: (a) First input $x$ is fixed. Therefore, all people in output images are the same person. (b) Second input $y$ is fixed. Therefore, all outputs have the same pose, but the people are different. All outputs are generated from only one generator $G$.}

\label{fig:3input_picture}
\vspace*{-3mm}
\end{figure*}

\textbf{The Effectiveness of FusionGAN:} Since previous works focused on set-to-set translation, they only work for two image sets. Therefore, to achieve transformation in three sets, ${}_{3}\!P_{2}=6$ generators were usually required. However, as mentioned above, we can transform images in multiple sets freely using only one generator. The generation results are shown in Figure~\ref{fig:3input_picture}. Figure~\ref{fig:3input_picture}\textcolor{red}{a} illustrates the case when the first input $x$ is fixed. Since the first input is fixed, the generated outputs have the same identity. The only thing that changes is the pose, which comes from the second images. Conversely, Figure~\ref{fig:3input_picture}\textcolor{red}{b} is the case when the second input $y$ is fixed. Since the second input is fixed, generated outputs have the same pose. However, the person is changed depending on the identity of the first input $x$. These results are generated from only one generator.

\begin{figure}[t]
\begin{center}
   \includegraphics[width=1.0\linewidth]{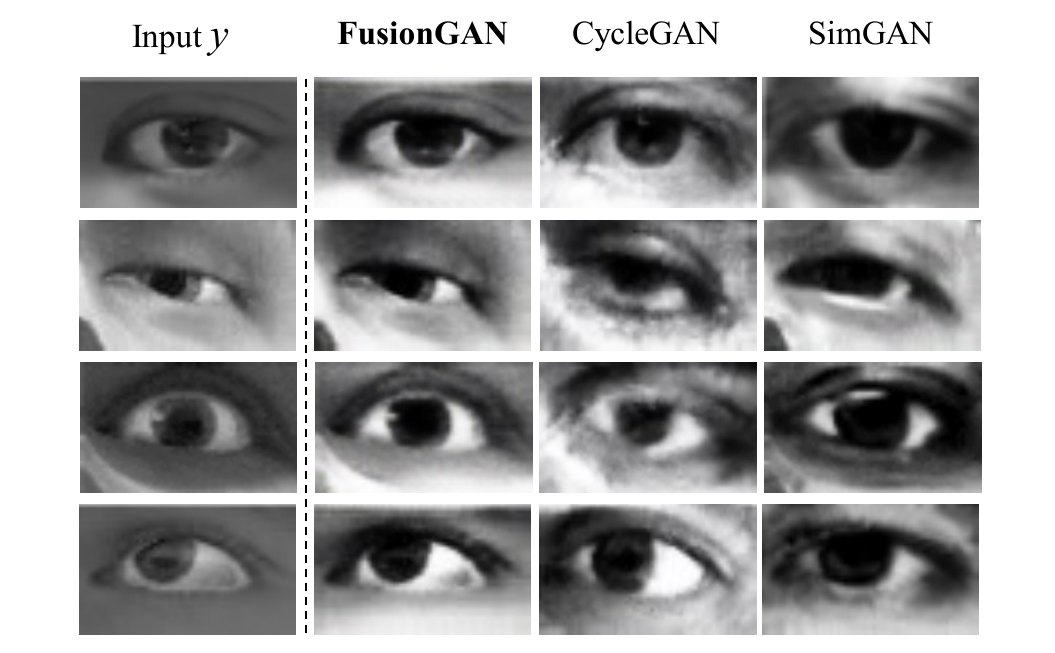}
\end{center}
\vspace*{-3mm}
   \caption{ Eye dataset results: from left to right, second input $y$, our method, CycleGAN~\cite{zhu2017unpaired}, and SimGAN~\cite{Shrivastava_2017_CVPR}. For our network, arbitrary real eye images were used for first input $x$. 
   \label{fig:eye}
\vspace*{-7mm}
}\
\end{figure}

\begin{figure*}[t]
\begin{center}
   \includegraphics[width=0.8\linewidth]{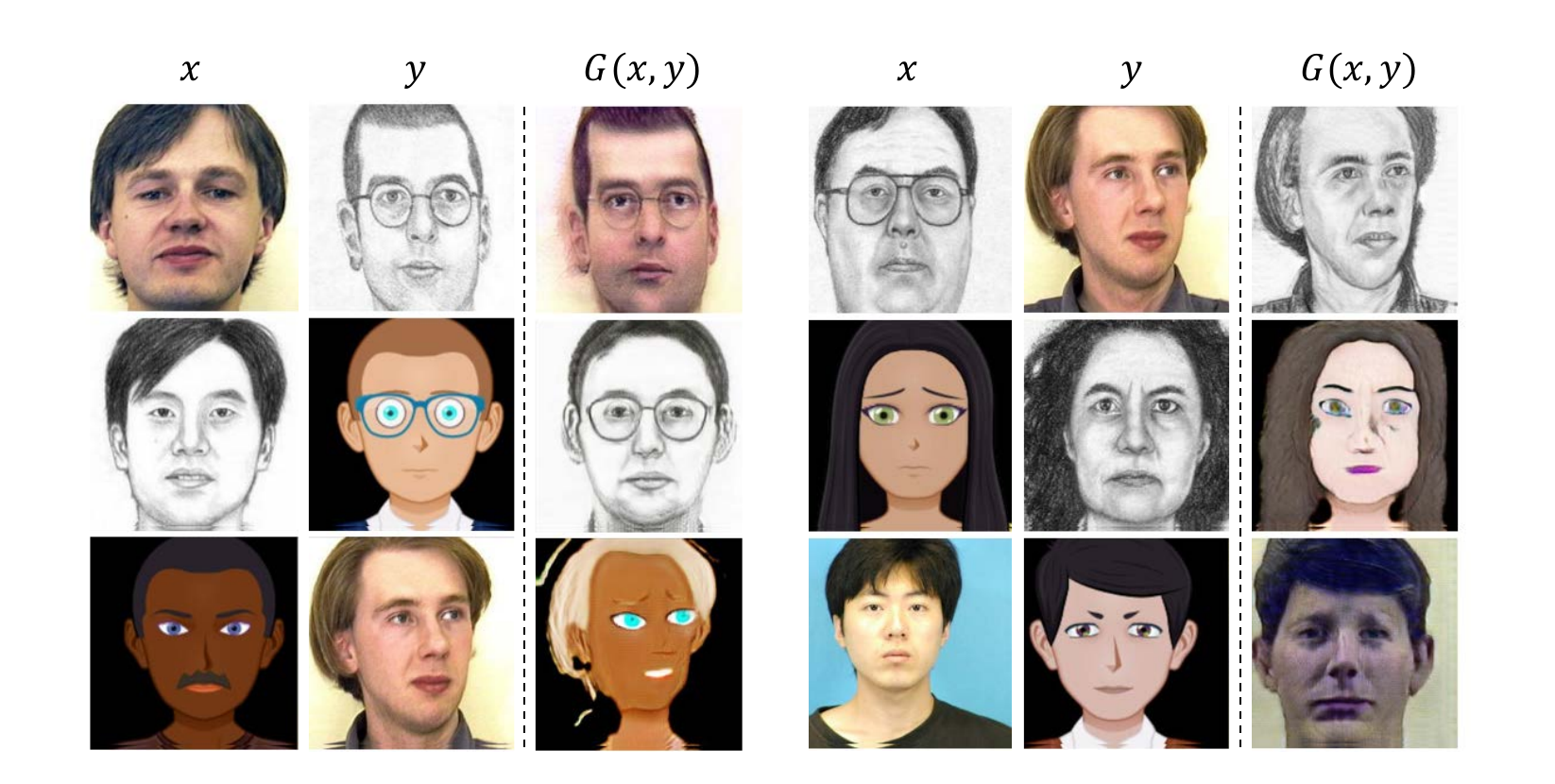}
\end{center}
\vspace*{-4mm}
   \caption{ Result images of FusionGAN in Photo--Sketch--Cartoon dataset. In this dataset, we define \textit{identity} as a style of the image and \textit{shape} as each person. When the network receives a photo as image $x$, then it produces photo style image of a human face in input $y$. All $6$ combinations of the dataset are shown.}
   \label{fig:cartoon}
\vspace*{-2mm}

\end{figure*}

\textbf{Quantitative Evaluation by Pose Detection:} Since the label for the pose is not given, we cannot directly estimate whether the output has desired pose. Instead, we can use existing body keypoint estimation that tells the location of each body joints to measure our quantitative results. We applied keypoint estimation on both second input image $y$ and generated output for three methods, and then computes the similarity to check whether $y$'s shape is well preserved. We applied pre-trained pose estimation model~\cite{Cao_2017_CVPR} to extract upper-body keypoints. To measure the difference of keypoints, we slightly modified existing measure object keypoint similarity (OKS)~\cite{MSCOCO}. While calculating OKS, the fixed penalty $100$ distance is applied when keypoint estimation has failed on detecting the existing keypoint of reference image ($256\times 256$ resolution setting). We experimented with various distance penalties and there was no significant difference. Total of randomly chosen $147$ images from $3$ sets were used for the estimation in Table~\ref{table:Youtube_posedetection}.

%

\begin{table}[H]
\small
	\begin{center}
		\begin{tabular}{|l|c|c |}
			\hline
			 & Keypoint similarity  \\
			\hline\hline
			FusionGAN & \textbf{0.52}  \\
			CycleGAN & 0.42 \\
			DiscoGAN & 0.38 \\
			\hline
		\end{tabular}
	\end{center}
	\caption{Keypoint similarity on YouTube Pose dataset}
\vspace*{-2mm}
	\label{table:Youtube_posedetection}
\end{table}
According to Table \ref{table:Youtube_posedetection}, our FusionGAN outperforms other baselines. It can be seen as that our proposed framework has generated an image that preserves the shape of target image better than the baselines.

\textbf{User Study: } To evaluate the quality of our generated images, we designed a
user study that asks three simple questions. \textbf{1)} Does generated image look like real person image? \textbf{2)} Is person in generated image the same person as input $x$?  \textbf{3)} Does the person in generated image have same pose as input $y$?
For this survey, each user was asked to score 24 source images and 72 images generated by three GAN methods. Users score the questions on a scale from 0 to 5 for each question and generated output. $32$ users participated in this survey.

Results on user study are shown in Table \ref{table: pose}. The subjects found that our generated outputs are realistic and well generated compare to the existing methods.

\vspace*{-1mm}
\begin{table}[H]
\small
	\begin{center}
		\begin{tabular}{|l|c|c|c |}
			\hline
			 & Reality & Identity & Pose  \\
			\hline\hline
			FusionGAN & \textbf{3.23} & \textbf{3.02} & \textbf{3.06}  \\
			CycleGAN & 2.77 & 2.59 & 2.79 \\
			DiscoGAN & 0.75 & 0.78 & 0.89 \\
			\hline
		\end{tabular}
	\end{center}
	\caption{User study on YouTube Pose Dataset}
	\label{table: pose}
\end{table}

\subsection{MPIIGaze and UnityEyes dataset}
\label{sec: eye}
SimGAN~\cite{Shrivastava_2017_CVPR} transformed simulated unreal eye images to realistic images using a GAN structure. They gathered many simulated eye images from UnityEyes~\cite{tan2002appearance} and trained a generator that can transform these simulated images into realistic images. Since real images are always limited, it is a great advantage to be able to create realistic images from synthetic images. By using this idea, SimGAN~\cite{Shrivastava_2017_CVPR} improved performance on the gaze estimation task. 

We performed the same tasks by using our FusionGAN. For this task, we prepared two image sets. Set $A$ has all simulated eye images from UnityEyes and set $B$ has all real eye images from the MPIIGaze dataset~\cite{zhang2015appearance}. Here, \textit{identity} refers to whether the image is real or simulated, and \textit{shape} refers to the other characteristics (gaze, wrinkle, eyelid, \etc) of each eye image. In this experiment, all the images have a resolution of $36\times 60$. For training, $13,939$ synthetic eye images from UnityEyes and $14,732$ real eye images from MPIIGaze dataset were used.

\textbf{Qualitative Results:} The result of our work for the eye datasets is shown in Figure~\ref{fig:eye}. For our network, we generated output $G(x,y)$ with some arbitrary real image $x$. CycleGAN~\cite{zhu2017unpaired} and SimGAN~\cite{Shrivastava_2017_CVPR} generate a realistic fake image with $y$ as input. Since this task is much easier than the one described in Section \ref{sec: YouTube}, all three networks generated realistic images. However, our network generates clearer images compare to the others.

Here, we need to note how well the shape is preserved. The outputs generated by SimGAN look very realistic, and the identities are translated well. However, if we compare the output images with the simulated image $y$, we see considerable deformation of the image details. For example, the gaze direction of the $4^{\text{th}}$ image is changed by both CycleGAN and SimGAN. 
\begin{figure*}[t]
\begin{center}
   \includegraphics[width=0.95\linewidth]{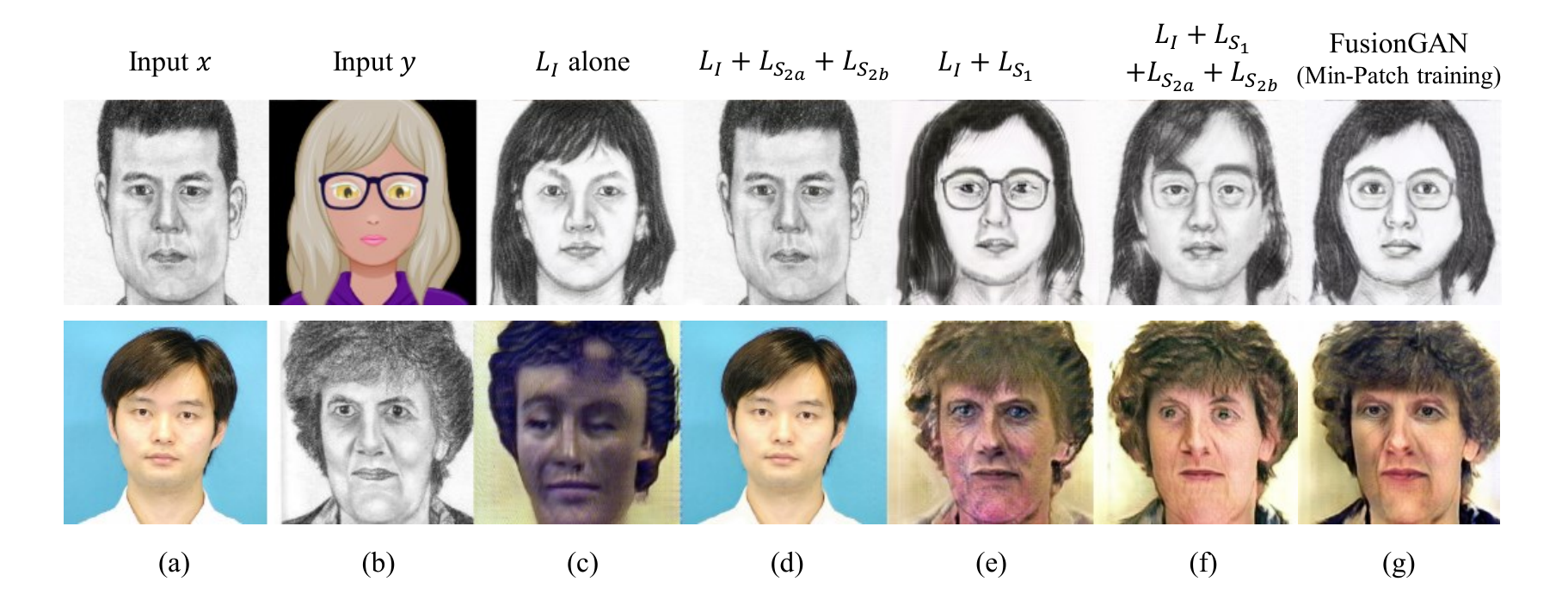}
\end{center}
\vspace*{-3mm}
   \caption{(Left) Ablation Study on our method. From left to right: input $x$, input $y$, $L_I$ alone, $L_I+L_{S_{2a}}+L_{S_{2b}}$, $L_I+L_{S_1}$, $L_I+L_{S_1}+L_{S_{2a}}+L_{S_{2b}}$, and FusionGAN which added Min-Patch training.}
\label{fig:ablation}
\vspace*{-5mm}
\end{figure*}

\textbf{User Study:} To evaluate the quality of our generated images, we designed a
user study that asks two simple questions. \textbf{1)} Is generated image realistic? \textbf{2)} Does generated eye image have same gaze direction as the reference image? For this survey, each user was asked to score 40 source images and 120 images generated by three GAN methods. $32$ users participated in this survey. 

\vspace*{-2mm}
\begin{table}[H]
\small
	\begin{center}
		\begin{tabular}{|l|c|c |}
			\hline
			 & Identity (Reality) & Shape (Eye gaze)  \\
			\hline\hline
			FusionGAN & \textbf{3.97} & \textbf{4.57}  \\
			CycleGAN & 2.55 & 2.63\\
			SimGAN & 3.02 & 3.57\\
			\hline
		\end{tabular}
	\end{center}
	\caption{User study on Eye Dataset}
	\label{table: eyegaze}
	\vspace*{-2mm}
\end{table}

Results on user study are shown in Table \ref{table: eyegaze}. The subjects found that our generated outputs are realistic and well generated compare to the existing methods. Especially, the ability to keep the shape of input $y$ is better than others.

\subsection{Photo--Sketch--Cartoon}
\label{sec:cartoon}
Lastly, we evaluated FusionGAN on the Photo--Sketch--Cartoon dataset, which consists of photographs, sketches, and cartoon images of people. Photographs were made by combining all $795$ images from the 2D face sets~\cite{PICS} and the CUHK student datasets~\cite{wang2009face}. A total of $560$ Sketch images were retrieved from the CUHK Face Sketch Database (CUFS)~\cite{wang2009face}. Cartoon image set was created by randomly combining attributes at cartoonify.com. We cropped $1,091$ images and resized them to $256\times 256$ pixels for training. As our definition of \textit{identity} and \textit{shape}, the image style is the identity for each set, and the shape was represented as who the person in an image is in the case of this dataset.

Figure~\ref{fig:cartoon} illustrates the result of our work with the Photo--Sketch--Cartoon dataset. FusionGAN produced successful results from this dataset. When a photography served as input $x$ and a sketch as input $y$, a realistic picture of sketch $y$ was created (left-top of Figure~\ref{fig:cartoon}). It is also possible to generate a cartoon image of a real face and realistic image of a cartoon.

\subsection{Ablation study}
We have suggested several novel ideas in this paper. Above experiments were conducted using all of the proposed methods. To investigate the strength of our new ideas, we designed an ablation study that shows how the performance changes depending on the presence or absence of each loss. Results are shown in Figure~\ref{fig:ablation}.

For (c), a random image is often generated, and the learning itself is also unstable. For the second row example, (c) does not generate realistic images because of unstable training. For (d), it is obvious that without $L_{S_1}$, the network pick the first image. We see that generated output is more like a second input $y$ after the shape loss $L_{S_1}$ is applied. In the first row example, only (e), (f) and (g) outputs the glasses following the $y$. Actually, (e) and (f) do not show the noticeable difference. However, because of the $L_{S_{2a}}+L_{S_{2b}}$, the outputs of (f) are slightly more like the style of $x$. After the Min-Patch training, FusionGAN shows more realistic outputs compared to the other methods. We notice that the unnatural parts of the output images are clearly reduced. Now it becomes more like the image what we aimed.

\section{Conclusion}

In this paper, we propose a new network, FusionGAN, that combines two input images by using identity loss $L_I$ and shape loss $L_S$. We have verified FusionGAN's ability to generate images that follow the identity of one image with the shape of another. FusionGAN can be applied to multiple sets and generate a reasonable image even if there is no ground truth for the desired result image. Using a subset of the YouTube Pose dataset, we changed a person's pose. In Eye dataset, we  transformed the synthetic eye image into a real image while maintaining the shape. We also applied our concept to show that Photo--Sketch--Cartoon character translation is possible with one model. 
\\\\{\large \textbf{Acknowledgement}}\\\\
This work was partly supported by the ICT R\&D program of MSIP/IITP, 2016-0-00563, Research on Adaptive Machine Learning Technology Development for Intelligent Autonomous Digital Companion.

{\small
\bibliographystyle{ieee}
\bibliography{egbib}

\begin{thebibliography}{10}\itemsep=-1pt

\bibitem{MSCOCO}
Mscoco keypoint evaluation metric.
\newblock \url{http://mscoco. org/dataset/\#keypoints-eval}.

\bibitem{PICS}
Psychological image collection at stirling (pics).
\newblock \url{http://pics.stir.ac.uk/}.

\bibitem{Cao_2017_CVPR}
Z.~Cao, T.~Simon, S.-E. Wei, and Y.~Sheikh.
\newblock Realtime multi-person 2d pose estimation using part affinity fields.
\newblock In {\em The IEEE Conference on Computer Vision and Pattern
  Recognition (CVPR)}, July 2017.

\bibitem{Charles16}
J.~Charles, T.~Pfister, D.~Magee, D.~Hogg, and A.~Zisserman.
\newblock Personalizing human video pose estimation.
\newblock In {\em IEEE Conference on Computer Vision and Pattern Recognition},
  2016.

\bibitem{Chen2016InfoGANIR}
X.~Chen, Y.~Duan, R.~Houthooft, J.~Schulman, I.~Sutskever, and P.~Abbeel.
\newblock Infogan: Interpretable representation learning by information
  maximizing generative adversarial nets.
\newblock In {\em NIPS}, 2016.

\bibitem{goodfellow2014generative}
I.~Goodfellow, J.~Pouget-Abadie, M.~Mirza, B.~Xu, D.~Warde-Farley, S.~Ozair,
  A.~Courville, and Y.~Bengio.
\newblock Generative adversarial nets.
\newblock In {\em Advances in neural information processing systems}, pages
  2672--2680, 2014.

\bibitem{Isola_2017_CVPR}
P.~Isola, J.-Y. Zhu, T.~Zhou, and A.~A. Efros.
\newblock Image-to-image translation with conditional adversarial networks.
\newblock In {\em The IEEE Conference on Computer Vision and Pattern
  Recognition (CVPR)}, July 2017.

\bibitem{pmlr-v70-kim17a}
T.~Kim, M.~Cha, H.~Kim, J.~K. Lee, and J.~Kim.
\newblock Learning to discover cross-domain relations with generative
  adversarial networks.
\newblock In {\em Proceedings of the 34th International Conference on Machine
  Learning}, pages 1857--1865, 2017.

\bibitem{kingma2013auto}
D.~P. Kingma and M.~Welling.
\newblock Auto-encoding variational bayes.
\newblock {\em arXiv preprint arXiv:1312.6114}, 2013.

\bibitem{Ledig_2017_CVPR}
C.~Ledig, L.~Theis, F.~Huszar, J.~Caballero, A.~Cunningham, A.~Acosta,
  A.~Aitken, A.~Tejani, J.~Totz, Z.~Wang, and W.~Shi.
\newblock Photo-realistic single image super-resolution using a generative
  adversarial network.
\newblock In {\em The IEEE Conference on Computer Vision and Pattern
  Recognition (CVPR)}, July 2017.

\bibitem{li2016precomputed}
C.~Li and M.~Wand.
\newblock Precomputed real-time texture synthesis with markovian generative
  adversarial networks.
\newblock In {\em European Conference on Computer Vision}, pages 702--716.
  Springer, 2016.

\bibitem{li2017generative}
Y.~Li, S.~Liu, J.~Yang, and M.-H. Yang.
\newblock Generative face completion.
\newblock In {\em Proceedings of the IEEE Conference on Computer Vision and
  Pattern Recognition}, volume~1, page~6, 2017.

\bibitem{ma2017pose}
L.~Ma, X.~Jia, Q.~Sun, B.~Schiele, T.~Tuytelaars, and L.~Van~Gool.
\newblock Pose guided person image generation.
\newblock {\em arXiv preprint arXiv:1705.09368}, 2017.

\bibitem{Radford2015UnsupervisedRL}
A.~Radford, L.~Metz, and S.~Chintala.
\newblock Unsupervised representation learning with deep convolutional
  generative adversarial networks.
\newblock {\em CoRR}, abs/1511.06434, 2015.

\bibitem{reed2016generative}
S.~Reed, Z.~Akata, X.~Yan, L.~Logeswaran, B.~Schiele, and H.~Lee.
\newblock Generative adversarial text to image synthesis.
\newblock In {\em International Conference on Machine Learning}, pages
  1060--1069, 2016.

\bibitem{reed2015deep}
S.~E. Reed, Y.~Zhang, Y.~Zhang, and H.~Lee.
\newblock Deep visual analogy-making.
\newblock In {\em Advances in neural information processing systems}, pages
  1252--1260, 2015.

\bibitem{ronneberger2015u}
O.~Ronneberger, P.~Fischer, and T.~Brox.
\newblock U-net: Convolutional networks for biomedical image segmentation.
\newblock In {\em International Conference on Medical image computing and
  computer-assisted intervention}, pages 234--241. Springer, 2015.

\bibitem{Shrivastava_2017_CVPR}
A.~Shrivastava, T.~Pfister, O.~Tuzel, J.~Susskind, W.~Wang, and R.~Webb.
\newblock Learning from simulated and unsupervised images through adversarial
  training.
\newblock In {\em The IEEE Conference on Computer Vision and Pattern
  Recognition (CVPR)}, July 2017.

\bibitem{sixt2016rendergan}
L.~Sixt, B.~Wild, and T.~Landgraf.
\newblock Rendergan: Generating realistic labeled data.
\newblock {\em arXiv preprint arXiv:1611.01331}, 2016.

\bibitem{smolensky1986information}
P.~Smolensky.
\newblock Information processing in dynamical systems: Foundations of harmony
  theory.
\newblock Technical report, COLORADO UNIV AT BOULDER DEPT OF COMPUTER SCIENCE,
  1986.

\bibitem{tan2002appearance}
K.-H. Tan, D.~J. Kriegman, and N.~Ahuja.
\newblock Appearance-based eye gaze estimation.
\newblock In {\em Applications of Computer Vision, 2002.(WACV 2002).
  Proceedings. Sixth IEEE Workshop on}, pages 191--195. IEEE, 2002.

\bibitem{Upchurch_2017_CVPR}
P.~Upchurch, J.~Gardner, G.~Pleiss, R.~Pless, N.~Snavely, K.~Bala, and
  K.~Weinberger.
\newblock Deep feature interpolation for image content changes.
\newblock In {\em The IEEE Conference on Computer Vision and Pattern
  Recognition (CVPR)}, July 2017.

\bibitem{wang2009face}
X.~Wang and X.~Tang.
\newblock Face photo-sketch synthesis and recognition.
\newblock {\em IEEE Transactions on Pattern Analysis and Machine Intelligence},
  31(11):1955--1967, 2009.

\bibitem{Zhang_2017_ICCV}
H.~Zhang, T.~Xu, H.~Li, S.~Zhang, X.~Wang, X.~Huang, and D.~N. Metaxas.
\newblock Stackgan: Text to photo-realistic image synthesis with stacked
  generative adversarial networks.
\newblock In {\em The IEEE International Conference on Computer Vision (ICCV)},
  Oct 2017.

\bibitem{zhang2015appearance}
X.~Zhang, Y.~Sugano, M.~Fritz, and A.~Bulling.
\newblock Appearance-based gaze estimation in the wild.
\newblock In {\em Proceedings of the IEEE Conference on Computer Vision and
  Pattern Recognition}, pages 4511--4520, 2015.

\bibitem{zhu2017unpaired}
J.-Y. Zhu, T.~Park, P.~Isola, and A.~A. Efros.
\newblock Unpaired image-to-image translation using cycle-consistent
  adversarial networks.
\newblock {\em arXiv preprint arXiv:1703.10593}, 2017.

\end{thebibliography}
}

\end{document}